\documentclass[runningheads]{llncs}
\usepackage[T1]{fontenc}
\usepackage{lmodern}
\usepackage{amsmath,amssymb,booktabs,graphicx,xcolor}
\usepackage{tikz}
\begin{document}
\title{FedPref: Federated Preference Learning for Structured Radiology Report Extraction}
\begingroup
\renewcommand{\thefootnote}{*}
\footnotetext{Preprint. Accepted at the 2nd Emerging LLM/LMM Applications in Medical Imaging (ELAMI 2026), held in conjunction with MICCAI 2026. To appear in the Springer proceedings.}
\endgroup
% uncomment below for including orcid
% \author{Flint Xiaofeng Fan\inst{1,2}\orcidID{0000-0003-1658-4699} \and
% Cheston Tan\inst{2}\orcidID{0000-0003-1248-4906} \and
% Yew-Soon Ong\inst{2}\orcidID{0000-0002-4480-169X} \and
% Roger Wattenhofer\inst{1}\orcidID{0000-0002-6339-3134}}

\author{Flint Xiaofeng Fan\inst{1,2} \and
Cheston Tan\inst{2} \and
Yew-Soon Ong\inst{2} \and
Roger Wattenhofer\inst{1}}
% %
\authorrunning{F. X. Fan et al.}
% \institute{
% ETH Zurich, Zurich, Switzerland
% \email{\{xiafan,wattenhofer\}@ethz.ch}
% \and
% Agency for Science, Technology and Research (A*STAR), Singapore, Singapore\\
% \email{\{fanx,cheston-tan,ong_yew_soon\}@a-star.edu.sg}}

\institute{
ETH Zurich, Zurich, Switzerland\\
\email{\{xiafan,wattenhofer\}@ethz.ch}
\and
Agency for Science, Technology and Research (A*STAR), Singapore, Singapore\\
\email{\{fanx,cheston-tan,ong\_yew\_soon\}@a-star.edu.sg}}
\titlerunning{FedPref for Structured Radiology Extraction}

\maketitle              % typeset the header of the contribution
\begin{abstract}
Radiology reports describe findings and locations in free text, but downstream search and analysis require these relations in a fixed schema. Learning this extraction requires labels that are unevenly distributed across institutions: smaller hospitals have less local evidence, and pooling data may be infeasible. We introduce FedPref: frozen public language models propose alternative JSON extractions, local annotations rank them, and sites collaboratively train compact Qwen3-8B adapters while sharing only model updates. A heterogeneous teacher pool provides cross-model contrast when repeated single-model samples collapse. On development data from six simulated hospitals with unequal data volume and disease prevalence, FedPref improves client-mean F1 by 2.49 points and worst-site F1 by 9.10 points compared with training each site in isolation, with the largest gains at the sites holding the least data. Central training on the pooled preference-pair union is 2.66 points higher on client-mean F1. On a locked, 400-report manually validated gold test set, FedPref reaches 68.68 F1 and pooled training 71.67, preserving that same ordering. FedPref thus lets institutions with unequal, unpooled data benefit from collaboration without ever sharing reports or annotations.
\end{abstract}

\section{Introduction}
Radiology reports encode two linked facts: which findings are present and where they occur. A report describing a left pleural effusion should yield the disease and its location; assigning the disease to the wrong anatomy remains an extraction error. Structuring these relations makes reports queryable and reusable as supervision. We study nine diseases and their locations in the centralized benchmark built from MIMIC-CXR and Chest ImaGenome~\cite{johnson2019mimiccxr,wu2021chest,sabour2025mind}.

Institutions observe unequal numbers and combinations of cases. Centralized training broadens coverage by pooling data; isolated training preserves institutional boundaries but leaves smaller hospitals with less evidence. Federated learning offers a third path by aggregating locally optimized model states~\cite{mcmahan2017communicationefficient,rieke2020future,NEURIPS2021_080acdcc,dai2024federated}. The schema and annotation rule stay fixed while sites differ in patient volume and disease prevalence.

\begin{figure*}[t]
\centering
\includegraphics[width=0.85\textwidth]{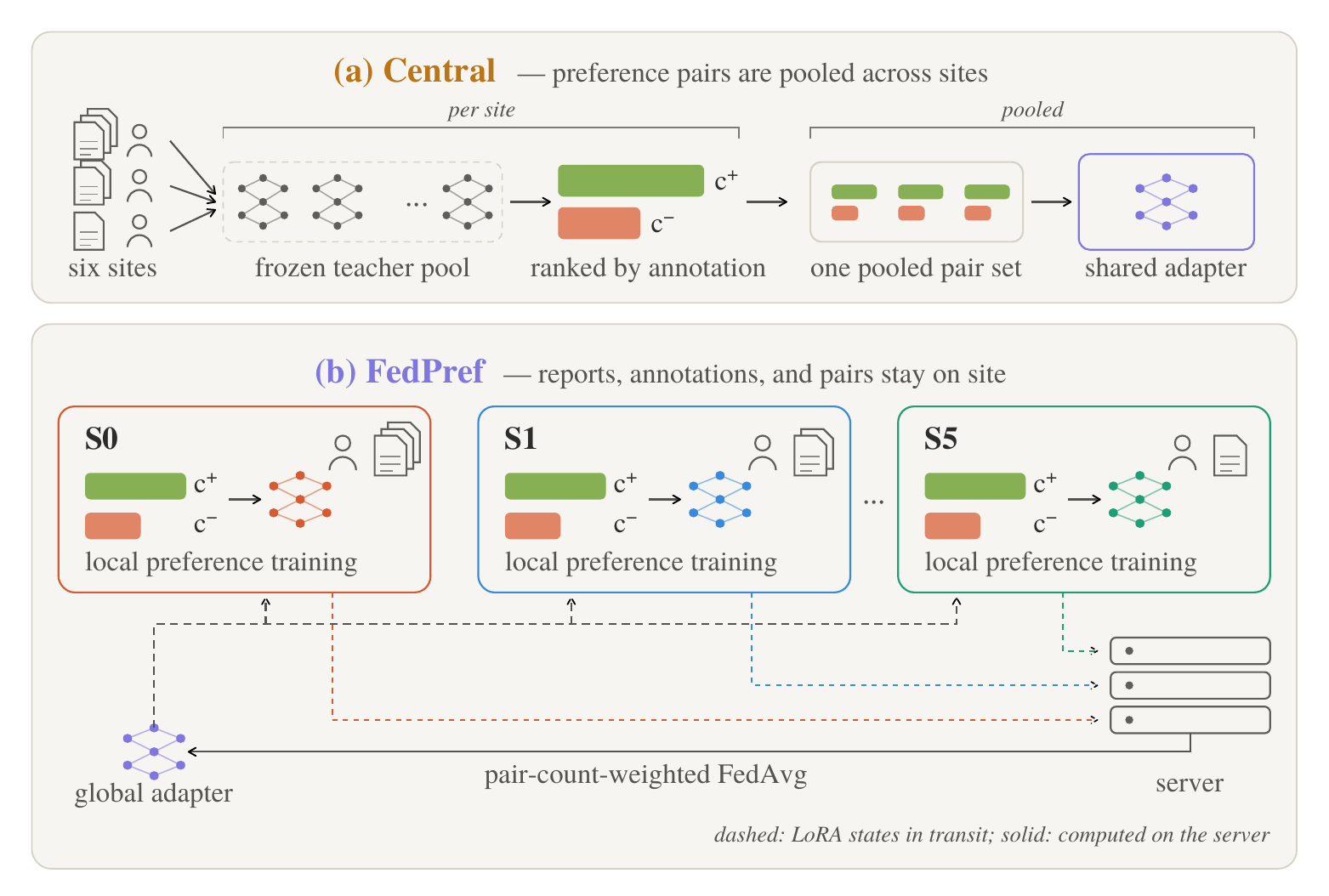}
\caption{FedPref intuition: pooled versus federated preference learning. At each site, frozen teachers propose alternative structured extractions and the local structured annotation ranks eligible candidates into chosen $c^+$ and rejected $c^-$. (a) Central pools the resulting preference pairs across sites and trains one shared adapter. (b) FedPref keeps reports, annotations, candidates, and preference pairs at their sites; each site performs local preference training, and only LoRA states are exchanged for pair-count-weighted FedAvg to form the global adapter. Local preference training denotes the chosen-response SFT initialization followed by fixed-reference DPO, detailed in Sec.~\ref{sec:method}.}
\label{fig:workflow}
\end{figure*}

FedPref turns structured annotations into a local preference interface (Fig.~\ref{fig:workflow}). For each report, four frozen teachers propose alternative JSON extractions; the site projects them into a common schema and uses its structured annotation to form a chosen response $c^+$ and rejected response $c^-$. Central provides the pooled counterpart by training on the union of retained preference pairs. FedPref instead keeps reports, annotations, candidates, and preference pairs local: sites train adapters locally and exchange only LoRA states, which pair-count-weighted Federated Averaging (FedAvg) combines into a shared adapter. 

Preference learning first requires rankable candidate contrast. We therefore run a separate feasibility diagnostic, outside the operational training pipeline, that tests repeated self-sampling from the Qwen3-8B target model. These samples often collapse to the same structured output, whereas heterogeneous cross-model candidates provide substantially more contrast (Sec.~\ref{sec:results}). This motivates the operational four-teacher pool and gives our first question: whether cross-model candidates supply useful local preference supervision. 

Each arm begins with one chosen-response supervised fine-tuning (SFT) epoch, followed by fixed-reference Direct Preference Optimization (DPO) ~\cite{rafailov2023direct}. Our second question is whether aggregating local preference updates improves client-mean and worst-site performance over isolated training, and how much of the performance attainable under pooled training it retains. We therefore evaluate three matched-exposure arms: Local trains each site in isolation, FedPref aggregates site adapters, and Central trains on the exact preference-pair union. On development data, FedPref improves client-mean and worst-site extraction over Local, with the largest gains at the two smallest sites, while Central measures the observed pooled--federated gap. A locked manually validated gold test checks generalization beyond automatically derived labels.

\section{Related Work}
MIMIC-CXR and Chest ImaGenome provide report-level and anatomy-linked supervision, while Sabour et al. formulate the centralized structured-extraction task used here~\cite{johnson2019mimiccxr,wu2021chest,sabour2025mind}.

Federated learning coordinates models without centralizing raw institutional records, making it attractive for medical data silos~\cite{mcmahan2017communicationefficient,rieke2020future}. Data locality alone does not guarantee privacy: formal protection requires a threat model and mechanisms such as secure aggregation or differential privacy~\cite{fan2025position}. FedAvg combines local states, while LoRA reduces the number of trainable parameters communicated~\cite{mcmahan2017communicationefficient,hu2022lora}. Factorwise LoRA averaging can nevertheless introduce an optimization gap~\cite{singhal2024fedex}, and unequal client utility motivates reporting both average and lower-tail performance~\cite{li2020fair}.

DPO learns directly from response rankings, and federated variants distribute feedback-driven, personalized, or medical-dialogue alignment~\cite{rafailov2023direct,fan2025fedrlhf,ma2025pffpo}. FedPref connects these strands by using local disease-location annotations to rank heterogeneous teacher outputs and aggregating only the resulting LoRA states.

\section{Method}\label{sec:method}
\paragraph{Overview.}
Site $k$ holds report-annotation pairs $\mathcal S_k=\{(x_i,y_i)\}$. For report $x_i$, teachers propose structures and annotation $y_i$ designates the best and worst eligible outputs as chosen response $c_i^+$ and rejected response $c_i^-$. Supervised fine-tuning (SFT) on chosen responses establishes the schema and fixed reference; DPO then refines which valid structures the model favors. FedPref aggregates LoRA states, Central trains on the pair union, and Local trains independently.

\paragraph{Local preference construction.}
Four frozen teachers, Qwen3-14B, Qwen2.5-14B-Instruct, Llama-3.1-8B-Instruct, and Mistral-7B-Instruct-v0.3, each generate one completion; the Qwen3-8B target is excluded~\cite{yang2025qwen}. Teachers receive the same task messages through native chat templates and use deterministic greedy decoding with reasoning disabled. Each completion is projected into the ontology and canonical JSON; failed projections are discarded. Eligible candidates are scored by
\begin{equation}
s(c,y)=0.70F_{1}^{\mathrm{loc+dis}}(c,y)+0.25F_{1}^{\mathrm{dis}}(c,y)+0.05\mathbb{1}[c\text{ valid}].
\label{eq:score}
\end{equation}
For an empty target set, we replace F1 by $1/(1+m)$ for $m$ predicted positives.
The larger weight on Location+Disease F1 reflects the primary task: a disease assigned to the wrong anatomy should not be preferred. Because only successfully projected candidates are eligible, the validity term records schema compliance without changing their ordering. The highest- and lowest-scoring candidates form $(c_i^+,c_i^-)$ when their structures differ, their margin is at least $0.10$, and $c_i^+$ has higher $F_{1}^{\mathrm{loc+dis}}$. The annotation determines ordering but never enters a teacher prompt or candidate.

\paragraph{Fixed-reference preference optimization.}
Round 0 performs one chosen-response SFT epoch. Central trains on the exact pair union, Local retains one root per site, and pair-weighted aggregation of those same local roots initializes FedPref. In DPO rounds $t=1,2,3$, each policy continues from its preceding checkpoint while its round-0 root remains the fixed reference $\pi_{\mathrm{ref}}$. Writing $h_\theta(c,x)=\log\pi_\theta(c\mid x)-\log\pi_{\mathrm{ref}}(c\mid x)$, site $k$ minimizes
\begin{equation}
\mathcal L_k(\theta)=-\mathbb E_{\mathcal D_k}\log\sigma\{\beta[h_\theta(c^+,x)-h_\theta(c^-,x)]\}.
\label{eq:dpo}
\end{equation}
The objective raises the chosen response's relative likelihood while the fixed SFT reference keeps that change comparable across rounds.

\paragraph{Aggregation and matched controls.}
At round $t$, every FedPref site receives the preceding global LoRA state $\phi^{(t-1)}$ and trains for one local epoch. With $n_k=|\mathcal D_k|$, the server computes
\begin{equation}
\phi^{(t)}=\sum_k\frac{n_k}{\sum_j n_j}\phi_k^{(t)}.
\label{eq:fedavg}
\end{equation}
Pair weighting makes each site's influence proportional to its retained training evidence. Central trains one adapter for three epochs on $\cup_k\mathcal D_k$; Local trains six isolated adapters for three local epochs, matching pair exposure across arms. In FedPref, reports, annotations, candidates, and preference pairs remain at their simulated sites; only LoRA states cross the boundary.

\section{Experimental Setup}\label{sec:setup}
\paragraph{Task and cohorts.}
We align MIMIC-CXR reports with Chest ImaGenome annotations~\cite{johnson2019mimiccxr,wu2021chest}. Automatically derived labels provide 4,800 training and 600 development reports; the manually validated Chest ImaGenome gold standard provides 100 diagnostic and 400 locked-test reports. The schema contains nine diseases and 18 locations, 15 with positive support. All cohorts are patient-disjoint; gold-standard patients were excluded before constructing the automatically labeled cohorts.
Code and reproduction instructions are available at \url{https://github.com/flint-xf-fan/FedPref}.

\paragraph{Simulated federation.}
We partition the data into six sites spanning 384 to 1,344 training reports and 297 to 717 retained pairs (Table~\ref{tab:siteeffects}). A capacity-constrained multilabel Dirichlet allocator ($\alpha=0.5$) varies disease prevalence while preserving site sizes. Prevalence divergence exceeds a capacity-matched IID allocation: 8.83 versus 1.89 points for training pairs and 10.14 versus 4.14 for development (both $p=0.0005$). The federation therefore combines unequal evidence with verified case-mix heterogeneity.

\paragraph{Preference data and optimization.}
The teachers produce 19,200 training and 2,400 development responses, yielding 2,961 training and 376 development pairs. All arms adapt Qwen3-8B with LoRA rank 16 and $\alpha=32$. SFT uses learning rate $2\times10^{-4}$; DPO uses $5\times10^{-6}$, $\beta=0.1$, BF16, cosine scheduling, and length 2,048. Each of three seeds independently trains the SFT root and all three DPO rounds, with three DPO exposures per pair in every arm.

\paragraph{Evaluation.}
Predictions are canonically projected before scoring; projection success records whether they contain a valid projectable JSON object. The primary endpoint is micro-F1 over positive disease-location labels (Location+Disease F1), which requires both finding and anatomy to be correct. Disease F1, projection success, false positives on reports with no targets, and held-out preference accuracy are secondary. Global F1 pools reports and weights larger sites more; client mean weights sites equally; worst-site F1 records the minimum. Central and FedPref use one shared policy, while home-routed Local sends each site's reports to its adapter. A paired patient-cluster bootstrap computes mean and worst-site effects over 10,000 draws and averages them across three seeds. Before gold-standard access, FedPref client-mean F1 selects one common round, breaking ties by worst-site F1 and then earlier round. Selected Central and FedPref policies are evaluated once on the locked gold-standard annotations.

\section{Results}\label{sec:results}
\paragraph{Cross-model candidates provide preference contrast.}
We measure candidate contrast by oracle headroom, the F1 gain from selecting the best candidate per report over the target baseline, and usable-pair coverage, the fraction of reports satisfying the pair margin. Across 100 automatically labeled reports, repeated Qwen3-8B sampling produced 71 identical parsed sets, 3.05 points of headroom, and 20.0\% coverage. A separate five-model diagnostic (four teachers plus the target) on 100 gold-standard reports produced 14.30 points and 76.0\%; its oracle exceeded the best individual model by 6.21 points (95\% CI, 3.70 to 9.01). The first diagnostic identifies target-model collapse; the second establishes cross-model contrast and motivates excluding the target from the operational four-teacher pool.

\begin{figure}[t]
\centering
\resizebox{0.90\linewidth}{!}{%
% Generated from verified multiround_trajectory.csv; do not edit manually.
\begin{tikzpicture}[x=1.05mm,y=1.05mm,line width=0.5pt]
\node[rotate=90, anchor=center, font=\scriptsize] at (2,20) {Location+Disease F1 (\%)};
\node[anchor=south, font=\small] at (27,36.3) {Client mean};
\draw (8,7) -- (46,7);
\draw (8,7) -- (8,33);
\draw[black!18] (8,7) -- (46,7);
\node[anchor=east, font=\tiny] at (7,7) {50};
\draw[black!18] (8,17.4) -- (46,17.4);
\node[anchor=east, font=\tiny] at (7,17.4) {60};
\draw[black!18] (8,27.8) -- (46,27.8);
\node[anchor=east, font=\tiny] at (7,27.8) {70};
\draw[black!18] (8,33) -- (46,33);
\node[anchor=east, font=\tiny] at (7,33) {75};
\draw (8,7) -- (8,6.3);
\node[anchor=north, font=\tiny] at (8,6) {0};
\draw (20.667,7) -- (20.667,6.3);
\node[anchor=north, font=\tiny] at (20.667,6) {1};
\draw (33.333,7) -- (33.333,6.3);
\node[anchor=north, font=\tiny] at (33.333,6) {2};
\draw (46,7) -- (46,6.3);
\node[anchor=north, font=\tiny] at (46,6) {3};
\node[anchor=north, font=\scriptsize] at (27,3.6) {Round};
\node[anchor=south, font=\small] at (77,36.3) {Worst site};
\draw (58,7) -- (96,7);
\draw (58,7) -- (58,33);
\draw[black!18] (58,7) -- (96,7);
\node[anchor=east, font=\tiny] at (57,7) {50};
\draw[black!18] (58,17.4) -- (96,17.4);
\node[anchor=east, font=\tiny] at (57,17.4) {60};
\draw[black!18] (58,27.8) -- (96,27.8);
\node[anchor=east, font=\tiny] at (57,27.8) {70};
\draw[black!18] (58,33) -- (96,33);
\node[anchor=east, font=\tiny] at (57,33) {75};
\draw (58,7) -- (58,6.3);
\node[anchor=north, font=\tiny] at (58,6) {0};
\draw (70.667,7) -- (70.667,6.3);
\node[anchor=north, font=\tiny] at (70.667,6) {1};
\draw (83.333,7) -- (83.333,6.3);
\node[anchor=north, font=\tiny] at (83.333,6) {2};
\draw (96,7) -- (96,6.3);
\node[anchor=north, font=\tiny] at (96,6) {3};
\node[anchor=north, font=\scriptsize] at (77,3.6) {Round};
\draw[blue!70!black] (8,15.01) -- (8,15.308);
\draw[blue!70!black] (7.45,15.01) -- (8.55,15.01);
\draw[blue!70!black] (7.45,15.308) -- (8.55,15.308);
\draw[blue!70!black] (20.667,27.045) -- (20.667,27.19);
\draw[blue!70!black] (20.117,27.045) -- (21.217,27.045);
\draw[blue!70!black] (20.117,27.19) -- (21.217,27.19);
\draw[blue!70!black] (33.333,28.944) -- (33.333,29.177);
\draw[blue!70!black] (32.783,28.944) -- (33.883,28.944);
\draw[blue!70!black] (32.783,29.177) -- (33.883,29.177);
\draw[blue!70!black] (46,28.961) -- (46,30.312);
\draw[blue!70!black] (45.45,28.961) -- (46.55,28.961);
\draw[blue!70!black] (45.45,30.312) -- (46.55,30.312);
\draw[blue!70!black] (8,15.159) -- (20.667,27.117);
\draw[blue!70!black] (20.667,27.117) -- (33.333,29.061);
\draw[blue!70!black] (33.333,29.061) -- (46,29.637);
\fill[blue!70!black] (8,15.159) circle (0.9);
\fill[blue!70!black] (20.667,27.117) circle (0.9);
\fill[blue!70!black] (33.333,29.061) circle (0.9);
\fill[blue!70!black] (46,29.637) circle (0.9);
\draw[red!75!black] (8,11.639) -- (8,12.622);
\draw[red!75!black] (7.45,11.639) -- (8.55,11.639);
\draw[red!75!black] (7.45,12.622) -- (8.55,12.622);
\draw[red!75!black] (20.667,18.378) -- (20.667,18.995);
\draw[red!75!black] (20.117,18.378) -- (21.217,18.378);
\draw[red!75!black] (20.117,18.995) -- (21.217,18.995);
\draw[red!75!black] (33.333,23.682) -- (33.333,24.757);
\draw[red!75!black] (32.783,23.682) -- (33.883,23.682);
\draw[red!75!black] (32.783,24.757) -- (33.883,24.757);
\draw[red!75!black] (46,26.692) -- (46,27.049);
\draw[red!75!black] (45.45,26.692) -- (46.55,26.692);
\draw[red!75!black] (45.45,27.049) -- (46.55,27.049);
\draw[red!75!black] (8,12.13) -- (20.667,18.686);
\draw[red!75!black] (20.667,18.686) -- (33.333,24.219);
\draw[red!75!black] (33.333,24.219) -- (46,26.87);
\fill[red!75!black] (8,13.345) -- (6.948,11.523) -- (9.052,11.523) -- cycle;
\fill[red!75!black] (20.667,19.901) -- (19.614,18.079) -- (21.719,18.079) -- cycle;
\fill[red!75!black] (33.333,25.434) -- (32.281,23.612) -- (34.386,23.612) -- cycle;
\fill[red!75!black] (46,28.085) -- (44.948,26.263) -- (47.052,26.263) -- cycle;
\draw[black!70] (8,12.571) -- (8,13.342);
\draw[black!70] (7.45,12.571) -- (8.55,12.571);
\draw[black!70] (7.45,13.342) -- (8.55,13.342);
\draw[black!70] (20.667,19.646) -- (20.667,19.918);
\draw[black!70] (20.117,19.646) -- (21.217,19.646);
\draw[black!70] (20.117,19.918) -- (21.217,19.918);
\draw[black!70] (33.333,22.899) -- (33.333,23.447);
\draw[black!70] (32.783,22.899) -- (33.883,22.899);
\draw[black!70] (32.783,23.447) -- (33.883,23.447);
\draw[black!70] (46,23.946) -- (46,24.615);
\draw[black!70] (45.45,23.946) -- (46.55,23.946);
\draw[black!70] (45.45,24.615) -- (46.55,24.615);
\draw[black!70] (8,12.957) -- (20.667,19.782);
\draw[black!70] (20.667,19.782) -- (33.333,23.173);
\draw[black!70] (33.333,23.173) -- (46,24.281);
\fill[black!70] (7.172,12.129) rectangle (8.828,13.785);
\fill[black!70] (19.839,18.954) rectangle (21.495,20.61);
\fill[black!70] (32.505,22.345) rectangle (34.161,24.001);
\fill[black!70] (45.172,23.453) rectangle (46.828,25.109);
\draw[blue!70!black] (58,12.837) -- (58,13.061);
\draw[blue!70!black] (57.45,12.837) -- (58.55,12.837);
\draw[blue!70!black] (57.45,13.061) -- (58.55,13.061);
\draw[blue!70!black] (70.667,24.757) -- (70.667,25.531);
\draw[blue!70!black] (70.117,24.757) -- (71.217,24.757);
\draw[blue!70!black] (70.117,25.531) -- (71.217,25.531);
\draw[blue!70!black] (83.333,26.888) -- (83.333,27.648);
\draw[blue!70!black] (82.783,26.888) -- (83.883,26.888);
\draw[blue!70!black] (82.783,27.648) -- (83.883,27.648);
\draw[blue!70!black] (96,26.636) -- (96,29.022);
\draw[blue!70!black] (95.45,26.636) -- (96.55,26.636);
\draw[blue!70!black] (95.45,29.022) -- (96.55,29.022);
\draw[blue!70!black] (58,12.949) -- (70.667,25.144);
\draw[blue!70!black] (70.667,25.144) -- (83.333,27.268);
\draw[blue!70!black] (83.333,27.268) -- (96,27.829);
\fill[blue!70!black] (58,12.949) circle (0.9);
\fill[blue!70!black] (70.667,25.144) circle (0.9);
\fill[blue!70!black] (83.333,27.268) circle (0.9);
\fill[blue!70!black] (96,27.829) circle (0.9);
\draw[red!75!black] (58,10.629) -- (58,11.009);
\draw[red!75!black] (57.45,10.629) -- (58.55,10.629);
\draw[red!75!black] (57.45,11.009) -- (58.55,11.009);
\draw[red!75!black] (70.667,16.06) -- (70.667,17.034);
\draw[red!75!black] (70.117,16.06) -- (71.217,16.06);
\draw[red!75!black] (70.117,17.034) -- (71.217,17.034);
\draw[red!75!black] (83.333,22.145) -- (83.333,23.121);
\draw[red!75!black] (82.783,22.145) -- (83.883,22.145);
\draw[red!75!black] (82.783,23.121) -- (83.883,23.121);
\draw[red!75!black] (96,25.619) -- (96,26.072);
\draw[red!75!black] (95.45,25.619) -- (96.55,25.619);
\draw[red!75!black] (95.45,26.072) -- (96.55,26.072);
\draw[red!75!black] (58,10.819) -- (70.667,16.547);
\draw[red!75!black] (70.667,16.547) -- (83.333,22.633);
\draw[red!75!black] (83.333,22.633) -- (96,25.845);
\fill[red!75!black] (58,12.034) -- (56.948,10.211) -- (59.052,10.211) -- cycle;
\fill[red!75!black] (70.667,17.762) -- (69.614,15.94) -- (71.719,15.94) -- cycle;
\fill[red!75!black] (83.333,23.848) -- (82.281,22.025) -- (84.386,22.025) -- cycle;
\fill[red!75!black] (96,27.06) -- (94.948,25.238) -- (97.052,25.238) -- cycle;
\draw[black!70] (58,9.095) -- (58,11.802);
\draw[black!70] (57.45,9.095) -- (58.55,9.095);
\draw[black!70] (57.45,11.802) -- (58.55,11.802);
\draw[black!70] (70.667,10.89) -- (70.667,11.01);
\draw[black!70] (70.117,10.89) -- (71.217,10.89);
\draw[black!70] (70.117,11.01) -- (71.217,11.01);
\draw[black!70] (83.333,11.668) -- (83.333,12.406);
\draw[black!70] (82.783,11.668) -- (83.883,11.668);
\draw[black!70] (82.783,12.406) -- (83.883,12.406);
\draw[black!70] (96,14.874) -- (96,17.89);
\draw[black!70] (95.45,14.874) -- (96.55,14.874);
\draw[black!70] (95.45,17.89) -- (96.55,17.89);
\draw[black!70] (58,10.449) -- (70.667,10.95);
\draw[black!70] (70.667,10.95) -- (83.333,12.037);
\draw[black!70] (83.333,12.037) -- (96,16.382);
\fill[black!70] (57.172,9.621) rectangle (58.828,11.277);
\fill[black!70] (69.839,10.122) rectangle (71.495,11.778);
\fill[black!70] (82.505,11.209) rectangle (84.161,12.865);
\fill[black!70] (95.172,15.554) rectangle (96.828,17.21);
\fill[blue!70!black] (22,-3.5) circle (0.855);
\node[anchor=west, font=\scriptsize] at (24.2,-3.5) {Central};
\fill[red!75!black] (52,-2.346) -- (51,-4.077) -- (53,-4.077) -- cycle;
\node[anchor=west, font=\scriptsize] at (54.2,-3.5) {FedPref};
\fill[black!70] (81.213,-4.287) rectangle (82.787,-2.713);
\node[anchor=west, font=\scriptsize] at (84.2,-3.5) {Local};
\end{tikzpicture}
% Markers: circle=Central, triangle=FedPref, square=Local. Error bars show three-seed sample standard deviation.
}
\caption{Performance across communication rounds under matched preference-pair exposure. Error bars show sample standard deviation over three seeds. FedPref overtakes home-routed Local on client mean at round 2 and retains a nine-point worst-site advantage at the selected third round.}
\label{fig:trajectory}
\end{figure}
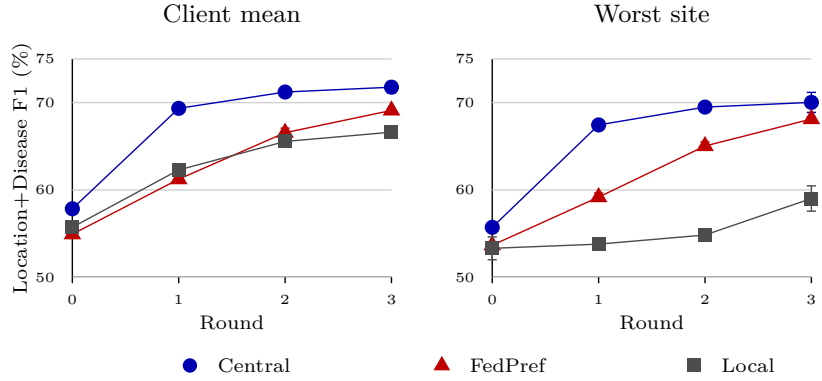

\paragraph{Federated aggregation improves isolated sites.}
FedPref improves across communication rounds (Fig.~\ref{fig:trajectory}). Its client-mean Location+Disease F1 increases from $54.93 \pm 0.47$ to $69.11 \pm 0.17$, while worst-site F1 increases from $53.67 \pm 0.18$ to $68.12 \pm 0.22$. It overtakes Local on client mean at round 2, and round 3 maximizes the prespecified FedPref client-mean selection criterion.

At the selected round (Table~\ref{tab:primary}), FedPref reaches $69.04 \pm 0.22$ global, $69.11 \pm 0.17$ client-mean, and $68.12 \pm 0.22$ worst-site F1. Home-routed Local reaches $66.62 \pm 0.32$ and $59.02 \pm 1.45$ on client mean and worst site. FedPref therefore improves the client mean by $+2.49$ points (95\% CI, $+1.24$ to $+3.79$) and the worst site by $+9.10$ points (95\% CI, $+1.01$ to $+11.87$).

Central reaches $71.77 \pm 0.67$, $71.77 \pm 0.65$, and $70.03 \pm 1.15$; Central exceeds FedPref by 2.66 points on client mean. FedPref and Local attain similar held-out preference accuracy, $91.49 \pm 0.27$ and $91.67 \pm 0.85$, despite their different extraction F1. Preference accuracy therefore does not track the extraction gap between the two systems. FedPref achieves $99.89 \pm 0.10\%$ projection success and has a $3.42 \pm 0.00\%$ false-positive rate on reports with no target findings.

\begin{table}[t]
\centering
\caption{Three-seed development performance at selected round 3. Values are mean $\pm$ sample standard deviation; paired effects are reported in text with patient-cluster confidence intervals.}
\label{tab:primary}
\small
\begin{tabular}{lrrrr}
\toprule
System & Global $F_{1}^{\mathrm{loc+dis}}$ & Mean $F_{1}^{\mathrm{loc+dis}}$ & Worst $F_{1}^{\mathrm{loc+dis}}$ & Pref. acc. \\
\midrule
Central & $71.77 \pm 0.67$ & $71.77 \pm 0.65$ & $70.03 \pm 1.15$ & $94.95 \pm 0.27$ \\
FedPref & $69.04 \pm 0.22$ & $69.11 \pm 0.17$ & $68.12 \pm 0.22$ & $91.49 \pm 0.27$ \\
Local & $67.32 \pm 0.28$ & $66.62 \pm 0.32$ & $59.02 \pm 1.45$ & $91.67 \pm 0.85$ \\
\bottomrule
\end{tabular}
\end{table}

\paragraph{Benefits concentrate at smaller sites.}
FedPref improves sites S1, S3, S4, and S5 (Table~\ref{tab:siteeffects}). The largest gains, $+9.46$ and $+4.47$ points, occur at S4 and S5, the two smallest training cohorts; S0 and S2 favor Local. In this partition, gains concentrate at sites with less local evidence.

\begin{table}[h!]
\centering
\caption{Per-site support (training reports / retained pairs) and selected-round FedPref--Local effect.}
\label{tab:siteeffects}
\small
\setlength{\tabcolsep}{3.0pt}
\begin{tabular}{@{}lrrrrrr@{}}
\toprule
Site & S0 & S1 & S2 & S3 & S4 & S5 \\
\midrule
Support & 1,344/717 & 1,056/591 & 864/556 & 672/486 & 480/297 & 384/314 \\
$\Delta F_{1}^{\mathrm{loc+dis}}$ (pt)
 & $-1.87$
 & $+0.87$
 & $-1.72$
 & $+3.73$
 & $+9.46$
 & $+4.47$ \\
\bottomrule
\end{tabular}
\end{table}

\paragraph{Locked gold-standard generalization.}
On the 400-report locked gold test, Central records Location+Disease F1 of $71.67 \pm 0.23$, Disease F1 of $90.12 \pm 0.17$, projection success of $99.58 \pm 0.29\%$, and a false-positive rate of $7.58 \pm 0.00\%$. FedPref records $68.68 \pm 0.35$, $86.36 \pm 0.16$, $100.00 \pm 0.00\%$, and $8.33 \pm 0.76\%$, respectively. The primary-endpoint comparison yields $+2.98$ points for Central minus FedPref (95\% patient-cluster CI, $+1.53$ to $+4.54$). Scores are mean $\pm$ sample standard deviation over three seeds, and the interval uses 10,000 patient-cluster resamples conditional on those seeds. Outputs from both policies project successfully for nearly every report and generalize to the gold-standard labels while preserving the Central--FedPref extraction ordering; the home-routed development results above provide the site-level evidence.

\section{Discussion and Conclusion}
FedPref reframes structured clinical annotations as a local preference interface. Each institution uses its existing annotations to rank candidate structures under a shared extraction rule, keeping reports and task-specific judgments under local control. Cross-model candidates supply local supervision that adapter aggregation carries into a shared model. This separation accommodates institutions that share a target schema but differ in data volume, case mix, and governance, and suggests a general strategy for federated adaptation of schema-constrained language models. The boundaries of the present evidence and the requirements for clinical deployment are detailed in Supplementary Sec.~A.1.

% In this controlled federation, aggregating adapters trained from local preference signals improves extraction most at the sites with the least evidence. Central training's remaining advantage measures the observed gap under matched exposure, and a locked gold test preserves the same ordering. FedPref thus offers a practical middle ground between isolated and pooled training for structured radiology extraction.

In this controlled federation, aggregating adapters trained from local preference signals improves extraction most at the sites with the least evidence. Central training's remaining advantage measures the observed pooled--federated gap under matched preference evidence and exposure, and a locked gold test preserves the same ordering. FedPref thus offers a practical middle ground between isolated and pooled training for structured radiology extraction.

\begin{credits}
% \subsubsection{\ackname} A bold run-in heading in small font size at the end of the paper is
% used for general acknowledgments, for example: This study was funded

\subsubsection{\discintname}
The authors have no competing interests to declare.
\end{credits}
\bibliographystyle{splncs04}
\bibliography{refs}




% LNCS Homepage, \url{http://www.springer.com/lncs}, last accessed 2023/10/25
% \end{thebibliography}

% ===== BEGIN SCIENTIFIC SUPPLEMENT =====
% Cut from this marker through the matching END marker to separate the supplement.
\clearpage
\appendix
\section{Scientific Supplement}

This supplement records the task contract, cohort construction, candidate
projection, federated optimization, and supporting results used in the main
paper. All tables report frozen experiment artifacts; no additional model
selection was performed for the supplement.

\subsection{Scope and Limitations}\label{sec:supp-scope}

FedPref is evaluated in a controlled six-site simulation derived from one
public dataset. The design varies site capacity and disease prevalence while
holding the ontology and annotation rule fixed. The evidence therefore does not
cover institution-specific reporting styles, annotation policies, clinical
workflows, or external domain shift; these questions require genuinely
independent institutional cohorts.

The locked test contains eight reports in three normalized-text duplicate
groups; one four-report group has inconsistent targets, introducing a small
source of evaluation ambiguity.

% The Central, FedPref, and Local arms use the same retained preference pairs and
% matched pair exposure. Their comparison estimates the effect of cross-site
% adapter aggregation within the chosen-response SFT plus fixed-reference DPO
% pipeline. The study does not include a matched Central/Federated/Local SFT
% control on canonical annotations, alternative preference objectives, or
% systematic sensitivity analyses over teacher composition, scoring weights,
% pair margin, DPO strength, and round count. The candidate-pool diagnostics
% establish available response contrast, but their distinct cohorts and candidate
% inventories preclude attributing downstream gains to teacher heterogeneity
% alone.

The Central, FedPref, and Local arms use the same retained preference-pair corpus and matched per-pair training exposure. FedPref and Local use the same client shards and share the same local chosen-response SFT roots; introducing cross-site aggregation gives FedPref its aggregated initialization and subsequent federated DPO updates. Their comparison therefore estimates the value of cross-site aggregation within this training pipeline. Central instead optimizes directly on the exact six-shard pair union and provides the pooled reference; Central--FedPref measures the observed pooled–federated gap under matched preference evidence and exposure, rather than an aggregation-only effect. The study does not systematically vary teacher composition, scoring weights, pair margin, DPO strength, or round count, nor does it compare alternative preference objectives or a matched canonical-annotation SFT baseline. The candidate-pool diagnostics establish available response contrast, but their distinct cohorts and candidate inventories preclude attributing downstream gains to teacher heterogeneity alone.

FedPref applies pair-count-weighted, factor-wise averaging to complete LoRA
states. Exact low-rank aggregation methods such as FedEx-LoRA are not compared;
factor-wise averaging need not equal averaging the clients' effective updates.
The protocol keeps reports and annotations local but provides no formal privacy
guarantee: communicated updates are not protected by secure aggregation or
differential privacy, and leakage attacks were not evaluated. Finally, the
six-client, all-participation, three-round study reports 7.81~GiB of cumulative
adapter traffic but does not establish runtime or network scaling under larger
or partially participating federations. Direct comparisons with other
federated preference methods remain future work.

\subsection{Task Contract and Cohort Support}

\paragraph{Output contract.}
The model receives a radiology report and returns exactly one JSON object with
two keys: \texttt{diseases}, a list of supported diseases, and
\texttt{findings}, a list of disease--location objects. The semantic prompt
applied the following rules to every target and teacher model:
\begin{enumerate}
    \item emit only the two required keys and no prose or Markdown;
    \item include positive, explicitly proposed, or hedged findings, while
    excluding findings that are only negated or absent;
    \item represent each localized finding as exactly one disease and one
    location from the frozen vocabularies in Table~\ref{tab:supp-vocab};
    \item include every localized disease in the disease list, while permitting
    a supported disease without a location when the report provides none;
    \item expand each right- or left-sided subregion to its corresponding
    whole-lung parent, without inferring a subregion from a whole-lung statement;
    \item deduplicate and sort diseases and findings in canonical vocabulary
    order; and
    \item emit empty lists when the report supports no target finding.
\end{enumerate}
The system message instructed the model to perform careful radiology extraction,
return one JSON object, use only the supplied vocabulary, and avoid unsupported
inference.

\begin{table}[!ht]
\centering
\caption{Frozen output vocabularies in canonical order.}
\label{tab:supp-vocab}
\small
\begin{tabular}{@{}p{0.42\linewidth}p{0.52\linewidth}@{}}
\toprule
Diseases & Locations \\
\midrule
lung opacity; pleural effusion; atelectasis; enlarged cardiac silhouette;
pulmonary edema/hazy opacity; pneumothorax; consolidation; fluid overload/heart
failure; pneumonia
& right lung; right apical zone; right upper lung zone; right mid lung zone;
right lower lung zone; right hilar structures; right costophrenic angle; left
lung; left apical zone; left upper lung zone; left mid lung zone; left lower
lung zone; left hilar structures; left costophrenic angle; mediastinum; upper
mediastinum; cardiac silhouette; trachea \\
\bottomrule
\end{tabular}
\end{table}

\paragraph{Cohort integrity.}
Table~\ref{tab:supp-location-support} reports location support for the
patient-disjoint cohorts defined in Sec.~4 of the main paper; gold-standard
patients were excluded from the automatically labeled training
and development data. Mediastinum, upper mediastinum, and trachea have zero
positive support because none of the nine target diseases localizes there in
these annotations; primary Location+Disease evaluation is therefore supported
over 15 regions.

\begin{table}[tb]
\centering
\caption{Report-level positive support by location. Training and development use
automatically derived labels; diagnostic and test use gold-standard labels.}
\label{tab:supp-location-support}
\small
\resizebox{\linewidth}{!}{%
\begin{tabular}{@{}lrrrr@{}}
\toprule
Location & Training ($n=4{,}800$) & Dev. ($n=600$) & Diagnostic ($n=100$) & Test ($n=400$) \\
\midrule
Right lung & 2,162 & 289 & 53 & 203 \\
Right apical zone & 173 & 26 & 3 & 15 \\
Right upper lung zone & 163 & 26 & 3 & 19 \\
Right mid lung zone & 416 & 57 & 8 & 29 \\
Right lower lung zone & 1,149 & 148 & 35 & 108 \\
Right hilar structures & 875 & 120 & 23 & 102 \\
Right costophrenic angle & 628 & 80 & 15 & 62 \\
Left lung & 2,263 & 304 & 54 & 206 \\
Left apical zone & 162 & 22 & 2 & 13 \\
Left upper lung zone & 128 & 14 & 3 & 10 \\
Left mid lung zone & 421 & 47 & 6 & 35 \\
Left lower lung zone & 1,362 & 168 & 32 & 122 \\
Left hilar structures & 847 & 120 & 26 & 97 \\
Left costophrenic angle & 703 & 92 & 14 & 62 \\
Mediastinum & 0 & 0 & 0 & 0 \\
Upper mediastinum & 0 & 0 & 0 & 0 \\
Cardiac silhouette & 950 & 111 & 27 & 101 \\
Trachea & 0 & 0 & 0 & 0 \\
\bottomrule
\end{tabular}}
\end{table}

\subsection{Candidate Generation, Projection, and Pair Construction}

\paragraph{Generation.}
Each frozen teacher generated one completion per report through its native chat
template. Decoding was greedy with a 768-token output limit, BF16 weights, and
no quantization or retrieval augmentation. Thinking was explicitly disabled
for Qwen3 models. Exact model revisions are recorded in the public reproduction
package\footnote{\url{https://github.com/flint-xf-fan/FedPref.}}.

\paragraph{Label-independent projection.}
Projection operated on teacher text without opening the reference annotation.
It extracted the first brace-balanced JSON object, rejected duplicate JSON keys
and nonstandard constants, and required both top-level lists. Extra top-level
fields were dropped and audited. Unknown vocabulary items and malformed finding
entries were dropped without synonym or fuzzy matching. Recognized finding
diseases were inserted into the disease list; subregions received their
whole-lung parents; and the result was deduplicated and canonically sorted. An
output with no valid object received the canonical empty fallback and was
ineligible for oracle or pair construction. Table~\ref{tab:supp-projection}
reports the resulting audit.

\begin{table}[tb]
\centering
\caption{Teacher-output projection audit. ``Strict'' denotes a valid raw JSON
object before projection; ``eligible'' denotes successful canonical projection.}
\label{tab:supp-projection}
\scriptsize
\resizebox{\linewidth}{!}{%
\begin{tabular}{@{}lrrrrrrrr@{}}
\toprule
& \multicolumn{5}{c}{Training ($4{,}800$ outputs per teacher)}
& \multicolumn{3}{c}{Development ($600$ per teacher)} \\
\cmidrule(lr){2-6}\cmidrule(l){7-9}
Teacher & Strict & Eligible & Discarded & Chosen & Rejected
& Strict & Eligible & Discarded \\
\midrule
Llama-3.1-8B & 3,650 & 4,771 & 29 & 815 & 1,219 & 440 & 598 & 2 \\
Mistral-7B & 3,375 & 4,672 & 128 & 606 & 1,350 & 414 & 583 & 17 \\
Qwen2.5-14B & 4,715 & 4,799 & 1 & 1,061 & 244 & 593 & 600 & 0 \\
Qwen3-14B & 4,697 & 4,800 & 0 & 479 & 148 & 590 & 600 & 0 \\
\bottomrule
\end{tabular}}
\end{table}

\paragraph{Preference-pair eligibility.}
For each report, the highest- and lowest-scoring projected teachers formed at
most one pair when their canonical structures differed, the composite-score
margin was at least 0.10, and the chosen candidate had strictly higher
Location+Disease F1. The training corpus retained 2,961 of 4,800 reports (61.69\%);
the development corpus retained 376 of 600 (62.67\%). Their mean score margins
were 0.426 and 0.414, with medians 0.400 and 0.382.

\subsection{Federation Construction and Non-IID Audit}

\paragraph{Capacity-constrained multilabel allocation.}
All reports were assigned to sites before pair scoring. Site capacities were
fixed at 28\%, 22\%, 18\%, 14\%, 10\%, and 8\% of each cohort. A multilabel
Dirichlet profile with $\alpha=0.5$ and skew weight 0.7 allocated intact report
rows. Development reused the label profiles derived from training, with independent
draw seeds 20260713 and 20260714. The realized site support is reported in
Table~\ref{tab:siteeffects} of the main paper.

\paragraph{Heterogeneity statistic.}
For site $k$, let $n_k$ be cohort size, $p_{k\ell}$ the prevalence of label
$\ell$, and $p_\ell$ its global prevalence. We used nine disease indicators
plus a normal-report indicator ($L=10$) and measured
\begin{equation}
D=\sqrt{\frac{1}{NL}\sum_k n_k\sum_{\ell=1}^{L}
       (p_{k\ell}-p_\ell)^2}, \qquad N=\sum_k n_k.
\end{equation}
The null distribution came from 1,999 capacity-preserving permutations of
intact multilabel report rows. The reported Monte Carlo probability uses the
plus-one correction. This audit was outcome-blind and completed before
training. Table~\ref{tab:supp-heterogeneity} reports all four cohorts; retained
training pairs and development reports are the training and evaluation cohorts
used for the primary study.

\begin{table}[tb]
\centering
\caption{Realized case-mix divergence against capacity-matched IID allocations.
Values of $D$ and the null 95th percentile are percentage points.}
\label{tab:supp-heterogeneity}
\small
\begin{tabular}{@{}lrrrrr@{}}
\toprule
Cohort & $N$ & Observed $D$ & Null 95th & Ratio & $p$ \\
\midrule
Training reports & 4,800 & 9.53 & 1.49 & 6.41 & 0.0005 \\
Retained training pairs & 2,961 & 8.83 & 1.89 & 4.67 & 0.0005 \\
Development reports & 600 & 10.14 & 4.14 & 2.45 & 0.0005 \\
Development retained pairs & 376 & 9.53 & 5.31 & 1.80 & 0.0005 \\
\bottomrule
\end{tabular}
\end{table}

Table~\ref{tab:supp-fit-prevalence} reports disease prevalence by site among
retained training pairs.

\begin{table}[tb]
\centering
\caption{Disease prevalence (\%) by site among retained training pairs. A report may
contain multiple diseases, so columns need not sum to 100.}
\label{tab:supp-fit-prevalence}
\scriptsize
\resizebox{\linewidth}{!}{%
\begin{tabular}{@{}lrrrrrrr@{}}
\toprule
Label & S0 & S1 & S2 & S3 & S4 & S5 & Global \\
\midrule
Normal & 34.3 & 43.1 & 23.6 & 7.2 & 24.2 & 6.7 & 25.7 \\
Atelectasis & 24.5 & 35.5 & 41.2 & 48.8 & 26.6 & 43.0 & 36.0 \\
Consolidation & 4.5 & 4.1 & 8.5 & 10.7 & 11.1 & 6.4 & 7.0 \\
Enlarged cardiac silhouette & 13.0 & 18.3 & 23.0 & 35.8 & 41.4 & 62.7 & 27.8 \\
Fluid overload/heart failure & 0.8 & 2.7 & 2.5 & 3.1 & 3.0 & 5.4 & 2.6 \\
Lung opacity & 61.5 & 50.8 & 73.4 & 89.5 & 62.6 & 72.3 & 67.4 \\
Pleural effusion & 15.8 & 15.6 & 21.4 & 35.2 & 29.0 & 27.1 & 22.5 \\
Pneumonia & 14.5 & 9.1 & 19.1 & 24.9 & 19.2 & 23.2 & 17.4 \\
Pneumothorax & 2.5 & 1.4 & 1.6 & 3.3 & 1.0 & 3.8 & 2.2 \\
Pulmonary edema/hazy opacity & 5.2 & 7.6 & 16.5 & 23.3 & 8.4 & 17.8 & 12.4 \\
\bottomrule
\end{tabular}}
\end{table}

\subsection{Optimization Details}

\paragraph{Adapter and training settings.}
All arms used LoRA rank 16 with $\alpha=32$, zero dropout, and no trainable bias, targeting the
\texttt{q}, \texttt{k}, \texttt{v}, \texttt{o}, \texttt{gate}, \texttt{up},
and \texttt{down} projections. BF16 training used length 2,048, fused AdamW,
cosine decay, 0.03 warmup, zero weight decay, gradient checkpointing, no packing
or shuffling, and fail-closed truncation. Physical batch size was one, with
gradient accumulation 24 for Central and four per client. SFT used one epoch at
$2\times10^{-4}$; each of three DPO rounds used one epoch at $5\times10^{-6}$,
$\beta=0.1$. Seeds were 20260713, 20260714, and 20260715.

\paragraph{Matched exposure and checkpoint lineage.}
Every pair received three DPO exposures in Central, FedPref, and Local. Round 1
continued from the corresponding round-0 SFT root; rounds 2 and 3 continued from
the preceding policy; FedPref clients received the previous global policy. The
SFT root remained the fixed DPO reference. Central opened the exact six-shard
union, while FedPref and Local used the same local shards and differed only by
aggregation. The server averaged complete LoRA states in float32 with pair-count
weights.

\subsection{Supporting Results}

\paragraph{Development trajectories.}
Table~\ref{tab:supp-trajectories} gives all prespecified communication rounds.
FedPref's client mean passes Local at round 2, while its worst-site advantage
widens from 0.36 points at round 0 to 9.10 points at round 3. The monotonic
FedPref trajectory supports the interpretation that cross-site evidence
accumulates through aggregation.

\begin{table}[!ht]
\centering
\caption{Development trajectories. Values are mean $\pm$ sample standard
deviation over three seeds; all metrics are percentages.}
\label{tab:supp-trajectories}
\small
\setlength{\tabcolsep}{4.0pt}
\begin{tabular}{@{}clrrr@{}}
\toprule
Round & System & Client-mean $F_1^{\mathrm{loc+dis}}$
& Worst-site $F_1^{\mathrm{loc+dis}}$ & Preference accuracy \\
\midrule
0 & Central & $57.85\pm0.14$ & $55.72\pm0.11$ & $86.70\pm0.27$ \\
  & FedPref & $54.93\pm0.47$ & $53.67\pm0.18$ & $82.00\pm0.41$ \\
  & Local & $55.73\pm0.37$ & $53.32\pm1.30$ & $83.95\pm0.15$ \\
\midrule
1 & Central & $69.34\pm0.07$ & $67.45\pm0.37$ & $90.51\pm0.15$ \\
  & FedPref & $61.24\pm0.30$ & $59.18\pm0.47$ & $86.44\pm0.27$ \\
  & Local & $62.29\pm0.13$ & $53.80\pm0.06$ & $87.85\pm0.31$ \\
\midrule
2 & Central & $71.21\pm0.11$ & $69.49\pm0.37$ & $94.15\pm0.00$ \\
  & FedPref & $66.56\pm0.52$ & $65.03\pm0.47$ & $89.63\pm0.00$ \\
  & Local & $65.55\pm0.26$ & $54.84\pm0.35$ & $89.98\pm0.31$ \\
\midrule
3 & Central & $71.77\pm0.65$ & $70.03\pm1.15$ & $94.95\pm0.27$ \\
  & FedPref & $69.11\pm0.17$ & $68.12\pm0.22$ & $91.49\pm0.27$ \\
  & Local & $66.62\pm0.32$ & $59.02\pm1.45$ & $91.67\pm0.85$ \\
\bottomrule
\end{tabular}
\end{table}

% ===== END SCIENTIFIC SUPPLEMENT =====

\end{document}